\pdfoutput=1
\documentclass[11pt,a4paper,table]{article}
\usepackage[hyperref]{acl2019}
\usepackage{times}
\usepackage{latexsym}
\usepackage{booktabs}
\usepackage{graphicx}
\usepackage{url}
\usepackage{xcolor}
\usepackage{pgf}
\usepackage{collcell}
\usepackage[T1]{fontenc}
\aclfinalcopy 
\def\aclpaperid{713} 

\newcommand{\ApplyGradient}[1]{%
  \pgfmathsetmacro{\PercentColor}{ifthenelse(#1-0 > 0, (#1-21)*.8, 0)}%
  \pgfmathsetmacro{\PercentInverse}{ifthenelse(\PercentColor > 70, 0, 100)}%
  \edef\x{\noexpand\cellcolor{red!\PercentColor}}\x\textcolor{black!\PercentInverse}{#1}%
}
\newcolumntype{R}{>{\collectcell\ApplyGradient}{c}<{\endcollectcell}}

\title{The Language of Legal and Illegal Activity on the Darknet}

\author{
	Leshem Choshen\thanks{\quad Equal contribution} \And Dan Eldad\footnotemark[1] \And Daniel Hershcovich\footnotemark[1]\\	
	School of Computer Science and Engineering \\
	The Hebrew University of Jerusalem \\
	\texttt{\{leshem.choshen,dan.eldad1,daniel.hershcovich,} \\
	\texttt{elior.sulem,omri.abend\}@mail.huji.ac.il}\\
	\And Elior Sulem\footnotemark[1]\And  Omri Abend \\
}
\date{}

\begin{document}
\maketitle

\begin{abstract}
  The non-indexed parts of the Internet (the Darknet)
   have become a haven for both legal and illegal anonymous activity.
  Given the magnitude of these networks, scalably monitoring their activity necessarily relies
    on automated tools, and notably on NLP tools.
  However, little is known about what characteristics texts communicated through the Darknet have, 
  and how well off-the-shelf NLP tools do on this domain.
  This paper tackles this gap and performs an in-depth investigation of the characteristics
    of legal and illegal text in the Darknet, comparing it to a clear net website with similar
    content as a control condition.
  Taking drug-related websites as a test case, we find that texts for selling legal and illegal drugs
    have several linguistic characteristics that distinguish them from one another, as well as from 
    the control condition, among them the distribution of POS tags, and the coverage of their named entities in Wikipedia.\footnote{Our code can be found in \url{https://github.com/huji-nlp/cyber}. Our data is available upon request.}
\end{abstract}

\section{Introduction}

  The term ``Darknet'' refers to the subset of Internet sites and pages that are not indexed by search engines.
  The Darknet is often associated with the ``.onion'' top-level domain, whose
  websites are referred to as ``Onion sites'', and are reachable via the Tor network anonymously.
  
  Under the cloak of anonymity, the Darknet harbors much illegal activity \citep{moore2016cryptopolitik}.
  Applying NLP tools to text from the Darknet is thus important for effective law enforcement and intelligence.
  However, little is known about the characteristics of the language used in the Darknet, 
  and specifically on what distinguishes text on websites that conduct legal and illegal activity.
	In fact, the only work we are aware of that classified Darknet texts into legal and illegal activity is \citet{Avarikioti18},
	but they too did not investigate in what ways these two classes differ.

  This paper addresses this gap, and studies the distinguishing features between legal and illegal texts in Onion sites,
  taking sites that advertise drugs as a test case. We compare our results to a control condition of texts 
  from eBay\footnote{\url{https://www.ebay.com}} pages that 
	advertise products corresponding to drug keywords.
 
  We find a number of distinguishing features. First, we confirm the results of \citet{Avarikioti18}, 
	that text from legal and illegal pages (henceforth, {\it legal} and {\it illegal} texts) can be distinguished based on the identity of the content words (bag-of-words) 
  in about 90\% accuracy over a balanced sample. Second, we find that the distribution of POS tags in the documents is a strong cue for 
	distinguishing between the classes (about 71\% accuracy). This indicates that the two classes are different in 
	terms of their syntactic structure. Third, we find that legal and illegal texts are roughly as distinguishable from one another as legal 
	texts and eBay pages are (both in terms of their words and their POS tags). 
	The latter point suggests that legal and illegal texts can be considered distinct domains, which explains why they can be 
	automatically classified, but also implies that applying NLP tools to Darknet texts is likely to face the obstacles of domain adaptation.  
  Indeed, we show that named entities in illegal pages are covered less well by Wikipedia, i.e., Wikification works less well on them.
  This suggests that for high-performance text understanding, specialized knowledge bases and tools may be needed for processing texts from the Darknet.
  By experimenting on a different domain in Tor (user-generated content), we show that the legal/illegal distinction generalizes across domains.

  After discussing previous works in Section \ref{sec:related_work}, we detail the datasets used in Section \ref{sec:datasets}. Differences in the vocabulary and named entities between the classes are analyzed in Section \ref{sec:domain}, before the presentation of the classification experiments (Section \ref{sec:classification}). Section \ref{sec:cross_domains} presents additional experiments, which explore cross-domain classification. We further analyze and discuss the findings in Section \ref{sec:discussion}.
  
  

%
%
%
%
%
%
%

\section{Related Work} \label{sec:related_work}

The detection of illegal activities on the Web is sometimes derived from a more general topic classification. For example, \citet{Biryukov14}
  classified the content of Tor hidden services into 18 topical categories, only some of which correlate with illegal activity.   
  \citet{GraczykKinningham15} combined unsupervised feature selection and an SVM classifier for the classification of drug sales in an anonymous marketplace.
  While these works classified Tor texts into classes, they did not directly address the legal/illegal distinction.


Some works directly addressed a specific type of illegality and a particular communication context. \citet{MorrisHirst12} used an SVM classifier to identify sexual predators in chatting message systems. The model includes both lexical features, including emoticons, and behavioral features that correspond to conversational patterns. Another example is the detection of pedophile activity in peer-to-peer networks \citep{Latapy13}, where a predefined list of keywords was used to detect child-pornography queries. Besides lexical features, we here consider other general linguistic properties, such as syntactic structure.

\citet{AlNabki17} presented DUTA (Darknet Usage Text Addresses), the first publicly available Darknet dataset, together with a manual classification into topical categories and sub-categories. For some of the categories, legal and illegal activities are distinguished. However, the automatic classification presented in their work focuses on the distinction between different classes of illegal activity, without addressing the distinction between legal and illegal ones, which is the subject of the present paper. \citet{AlNabki19} extended the dataset to form DUTA-10K, which we use here. Their results show that 20\% of the hidden services correspond to ``suspicious'' activities. The analysis was conducted using the text classifier presented in \citet{AlNabki17} and manual verification. 

Recently, \citet{Avarikioti18} presented another topic classification of text from Tor together with a first classification into legal and illegal activities.  The experiments were performed on a newly crawled corpus obtained by recursive search. The legal/illegal classification was done using an SVM classifier in an active learning setting with bag-of-words features. Legality was assessed in a conservative way where illegality is assigned if the purpose of the content is an obviously illegal action, even if the content might be technically legal. They found that a linear kernel worked best and reported an F1 score of 85\% and an accuracy of 89\%. Using the dataset of \citet{AlNabki19}, and focusing on specific topical categories, we here confirm the importance of content words in the classification, and explore the linguistic dimensions supporting classification into legal and illegal texts. 
\begin{table*}[]
    \resizebox{\textwidth}{!}{%
        \begin{tabular}{@{}ll|rrr|rrr|rrr|rrr@{}}
&& \multicolumn{3}{c|}{All Onion} & \multicolumn{3}{c|}{eBay}
& \multicolumn{3}{c|}{Illegal Onion} & \multicolumn{3}{c}{Legal Onion} \\
&& all & half 1 & half 2 & all & half 1 & half 2
& all & half 1 & half 2 & all & half 1 & half 2 \\
\hline
& all & & 0.23 & 0.25 & 0.60 & 0.61 & 0.61 & 0.33 & 0.39 & 0.41 & 0.35 & 0.41 & 0.42 \\
All Onion & half 1 & 0.23 & & 0.43 & 0.60 & 0.62 & 0.62 & 0.37 & 0.33 & 0.50& 0.40 & 0.36 & 0.52 \\
& half 2 & 0.25 & 0.43 & & 0.61 & 0.62 & 0.62 & 0.39 & 0.50& 0.35 & 0.39 & 0.51 & 0.35 \\
\hline
& all & 0.60 & 0.60 & 0.61 & & 0.23 & 0.25 & 0.59 & 0.60& 0.60& 0.66 & 0.67 & 0.67 \\
eBay & half 1 & 0.61 & 0.62 & 0.62 & 0.23 & & 0.43 & 0.60 & 0.61 & 0.61 & 0.67 & 0.67 & 0.68 \\
& half 2 & 0.61 & 0.62 & 0.62 & 0.25 & 0.43 & & 0.60 & 0.61 & 0.61 & 0.67 & 0.68 & 0.68 \\
\hline
& all & 0.33 & 0.37 & 0.39 & 0.59 & 0.60 & 0.60 & & 0.23 & 0.27 & 0.61 & 0.62 & 0.62 \\
Illegal Onion & half 1 & 0.39 & 0.33 & 0.50 & 0.60 & 0.61 & 0.61 & 0.23 & & 0.45 & 0.62 & 0.63 & 0.62 \\
& half 2 & 0.41 & 0.50 & 0.35 & 0.60 & 0.61 & 0.61 & 0.27 & 0.45 & & 0.62 & 0.63 & 0.63 \\
\hline
& all & 0.35 & 0.40 & 0.39 & 0.66 & 0.67 & 0.67 & 0.61 & 0.62 & 0.62 & & 0.26 & 0.26 \\
Legal onion & half 1 & 0.41 & 0.36 & 0.51 & 0.67 & 0.67 & 0.68 & 0.62 & 0.63 & 0.63 & 0.26 & & 0.47 \\
& half 2 & 0.42 & 0.52 & 0.35 & 0.67 & 0.68 & 0.68 & 0.62 & 0.62 & 0.63 & 0.26 & 0.47 &  \\          
        \end{tabular}%
    }
    \caption{Jensen-Shannon divergence between word distribution in all Onion drug sites, Legal and Illegal Onion drug sites, and eBay sites.
    Each domain was also split in half for within-domain comparison. \label{ta:domain_halves}}
\end{table*}

\section{Datasets Used}\label{sec:datasets}

\paragraph{Onion corpus.}
We experiment with data from Darknet websites containing
legal and illegal activity, all from the DUTA-10K corpus \citep{AlNabki19}.
We selected the ``drugs'' sub-domain as a test case, as it is a large domain in the corpus,
that has a ``legal'' and ``illegal'' sub-categories, and where the distinction between them 
can  be reliably made.
These websites advertise and sell
drugs, often to international customers.
While legal websites often sell pharmaceuticals,
illegal ones are often related to substance abuse.
These pages are directed by sellers to their customers.
  

\paragraph{eBay corpus.}
As an additional dataset of similar size and characteristics,
but from a clear net source, and of legal nature,
we compiled a corpus of eBay pages.
eBay is one of the largest hosting sites for retail sellers of various goods. Our corpus contains 118 item descriptions, each consisting of more than one sentence.
Item descriptions vary in price, item sold and seller. The descriptions were selected by searching eBay for drug related terms,\footnote{Namely,  \textit{marijuana}, \textit{weed}, \textit{grass} and \textit{drug}.} and selecting search patterns to avoid over-repetition. For example, where many sell the same product, only one example was added to the corpus. Search queries also included filtering for price, so that each query resulted with different items. Either because of advertisement strategies or the geographical dispersion of the  sellers, the eBay corpus contains formal as well as informal language, and some item descriptions contain abbreviations and slang.
Importantly, eBay websites are assumed to conduct legal activity---even
  when discussing drug-related material, we find it is never the sale of illegal
  drugs but rather merchandise, tools, or otherwise related content.

\paragraph{Cleaning.} 
As preprocessing for all experiments, we apply some cleaning to the text
of web pages in our corpora.
HTML markup is already removed in the original datasets,
but much non-linguistic content remains, such as
buttons, encryption keys, metadata and URLs.
We remove such text from the web pages, and join paragraphs 
to single lines (as newlines are sometimes present in the original dataset for display purposes only).
We then remove any duplicate paragraphs, where paragraphs are considered
identical if they share all but numbers (to avoid an over-representation of some remaining surrounding text from the websites, e.g. ``Showing all 9 results'').

\section{Domain Differences}\label{sec:domain}

As pointed out by \citet{Plank11}, there is no common ground as to what constitutes a domain. Domain differences are attributed in some works to differences in vocabulary \citep{Blitzer06} and in other works to differences in style, genre and medium \citep{McClosky2010}. While here we adopt an existing classification, based on the DUTA-10K corpus, 
we show in which way and to what extent it translates to distinct properties of the texts. 
This question bears on the possibility of distinguishing between legal and illegal drug-related websites based on their text alone (i.e., without recourse to additional information, such as meta-data or network structure).
    
We examine two types of domain differences between legal and illegal texts: vocabulary differences and named entities. 

\subsection{Vocabulary Differences}

    To quantify the domain differences between texts from legal and illegal texts,
    we compute the frequency distribution of words in the eBay corpus, the legal and illegal drugs Onion corpora, and the entire Onion drug section (All~Onion). 
    Since any two sets of texts are bound to show some disparity between them, we compare the differences between domains to a control setting, where we randomly split each examined corpus into two halves, and compute the frequency distribution of each of them.
    The inner consistency of each corpus, defined as the similarity of distributions between the two halves, serves as a reference point for the similarity between domains.
    We refer to this measure as ``self-distance''.
    
    Following \citet{Plank2011EffectiveMO}, we compute the Jensen-Shannon divergence and Variational distance (also known as L1 or Manhattan) as the comparison measures between the word frequency histograms.
    
    Table~\ref{ta:domain_halves} presents our results.
    The self-distance of the eBay, Legal Onion and Illegal Onion corpora lies between 0.40 to 0.45
    by the Jensen-Shannon divergence, but the distance between each pair is 0.60 to 0.65, with the three approximately forming an equilateral triangle
    in the space of word distributions.
    Similar results are obtained using Variational distance, and are omitted for brevity.

    These results suggest that rather than regarding all drug-related Onion texts as one domain, with legal and illegal texts as sub-domains, 
      they should be treated as distinct domains.
    Therefore, using Onion data to characterize the differences between illegal and legal linguistic attributes is sensible. 
    In fact, it is more sensible than comparing Illegal Onion to eBay text, as there the legal/illegal distinction may be confounded by the differences
      between eBay and Onion data.

\begin{table}
\begin{center}
\begin{tabular}{l|r}
 & \% Wikifiable\\
 \hline
eBay & $38.6 \pm2.00$\\
Illegal Onion & $32.5 \pm1.35$\\
Legal Onion & $50.8 \pm2.31$
\end{tabular}
\end{center}
\caption{Average percentage of wikifiable named entities in a website per domain, with standard error.\label{ta:wiki}}
\end{table}

\subsection{Differences in Named Entities}\label{sec:ner}

    In order to analyze the difference in the distribution of 
    named entities between the domains,  we used a Wikification technique \cite{bunescu2006using}, i.e., linking entities to their corresponding article in Wikipedia.

    Using spaCy's\footnote{\url{https://spacy.io}}
    named entity recognition, we first extract all named
    entity mentions from all the corpora.\footnote{We use all named entity types provided by spaCy (and not only ``Product'') to get a broader perspective on the differences between the domains in terms of their named entities. For example, the named entity ``Peru'' (of type ``Geopolitical Entity'') appears multiple times in Onion sites and is meant to imply the quality of a drug.} 
    We then search for relevant Wikipedia entries for each named entity using the DBpedia Ontology API \cite{isem2013daiber}.
    For each domain we compute the total number of named entities and the percentage with corresponding Wikipedia articles.

    The results were obtained by averaging the percentage of wikifiable named entities in each site per domain. We also report the standard error for each average.
    According to our results (Table~\ref{ta:wiki}), the Wikification success ratios of eBay and Illegal Onion named entities is comparable and relatively low. However, sites selling legal drugs on Onion have a much higher Wikification percentage.

Presumably the named entities in Onion sites selling legal drugs are
more easily found in public databases such as Wikipedia because they
are mainly well-known names for legal pharmaceuticals. However, in
both Illegal Onion and eBay sites, the list of named entities includes
many slang terms for illicit drugs and paraphernalia. These slang terms
are usually not well known by the general public, and are therefore
less likely to be covered by Wikipedia and similar public databases.

In addition to the differences in Wikification ratios between
the domains, it seems spaCy had trouble correctly identifying
named entities in both Onion and eBay sites, possibly
due to the common use of informal language and drug-related jargon.
Eyeballing the results, there were a fair number of false positives (words and phrases that were found by spaCy but were not actually named entities),
especially in Illegal Onion sites.
In particular, slang terms for drugs, as well as abbreviated drug terms, for example ``kush'' or ``GBL'',
were being falsely picked up by spaCy.\footnote{We
consider these to be false positives as they are names of substances but not specific products, 
and hence not named entities. See \url{https://spacy.io/api/annotation\#section-named-entities}}

To summarize, results suggest both that (1) legal and illegal texts are different in terms of their named entities and their coverage in Wikipedia, as well as that (2) standard databases and standard NLP tools for named entity recognition (and potentially other text understanding tasks), require considerable adaptation to be fully functional on text related to illegal activity.

\section{Classification Experiments} \label{sec:classification}

  Here we detail our experiments in classifying text from different legal and
  illegal domains using various methods, to find the most important linguistic features
  distinguishing between the domains. Another goal of the classification task is to confirm our finding that the domains are distinguishable.

\paragraph{Experimental setup.}

We split each subset among
\{eBay, Legal Onion, Illegal Onion\}
into training, validation and test.
We select 456 training paragraphs, 57 validation paragraphs and
58 test paragraphs for each category (approximately a 80\%/10\%/10\% split),
randomly downsampling larger categories for an even division of labels.

\paragraph{Model.}

To classify paragraphs into categories, we experiment with five classifiers:

\begin{itemize}
  \item NB (Naive Bayes) classifier
    with binary bag-of-words features, i.e., indicator feature for each word. This simple classifier
    features  frequently in work on text classification in the Darknet.
  \item SVM (support vector machine) classifier with an RBF kernel,
  also with BoW features that count the number of words of each type.
  \item BoE (bag-of-embeddings): each word is represented with its 100-dimensional
  GloVe vector \cite{pennington2014glove}. BoE$_\mathrm{sum}$ (BoE$_\mathrm{average}$) sums (averages) 
  the embeddings for all words in the paragraph to a single vector, and applies a 100-dimensional fully-connected layer with
  ReLU non-linearity and dropout $p=0.2$. The word vectors are not updated during training.
  Vectors for words not found in GloVe are set randomly
  $\sim\mathcal{N}(\mu_\textrm{GloVe},\sigma^2_\textrm{GloVe})$.
  \item seq2vec: same as BoE, but instead of averaging word vectors,
  we apply a single-layer 100-dimensional BiLSTM to the word vectors, and take the concatenated
  final hidden vectors from the forward and backward part as the input to a
  fully-connected layer (same hyper-parameters as above).
  \item attention: we replace the word representations with contextualized
  pre-trained representations from ELMo \cite{Peters:2018}. We then apply a self-attentive
  classification network \cite{mccann2017learned} over the contextualized representations. This architecture has proved very effective for classification in
  recent work \cite{W18-5427,D18-1401}.
\end{itemize}

For the NB classifier we use \texttt{BernoulliNB} from
\texttt{scikit-learn}\footnote{\url{https://scikit-learn.org}}
with $\alpha=1$,
and for the SVM classifier we use \texttt{SVC}, also from \texttt{scikit-learn},
with $\gamma=\mathrm{scale}$ and tolerance=$10^{-5}$.
We use the AllenNLP  library\footnote{\url{https://allennlp.org}}
\cite{Gardner2017AllenNLP} to implement the neural network classifiers.

\paragraph{Data manipulation.}

In order to isolate what factors contribute to the classifiers' performance,
we experiment with four manipulations to the input
data (in training, validation and testing).
Specifically, we examine the impact of variations in the content words, function words and shallow syntactic structure (represented through POS tags).
For this purpose, we consider content words as words whose universal part-of-speech
according to spaCy is one of the following:
\[\{\textsc{adj, adv, noun, propn, verb, x, num}\}\]
and function words as all other words.
The tested manipulations are:

\begin{itemize}
  \item Dropping all content words ({\it drop cont.})
  \item Dropping all function words ({\it drop func.})
  \item Replacing all content words with their universal part-of-speech ({\it pos cont.})
  \item Replacing all function words with their universal part-of-speech ({\it pos func.})
\end{itemize}

Results when applying these manipulations are compared to the {\it full} condition, where all
words are available.

\paragraph{Settings.}

We experiment with two settings, classifying paragraphs from different domains:
\begin{itemize}
  \item Training and testing on eBay pages vs. Legal drug-related Onion pages,
  as a control experiment
  to identify whether Onion pages differ from clear net pages.
  \item Training and testing on Legal Onion vs. Illegal Onion drugs-related pages,
  to identify the difference in language between legal and illegal activity
  on Onion drug-related websites.
  
\end{itemize}

\begin{table}[t]
\centering
\setlength\tabcolsep{4pt}
\begin{tabular}{l *{5}{R}}
&& \multicolumn{1}{c}{\bf drop} & \multicolumn{1}{c}{\bf drop} & \multicolumn{1}{c}{\bf pos} & \multicolumn{1}{c}{\bf pos}\\
& \multicolumn{1}{c}{\bf full} & \multicolumn{1}{c}{\bf cont.} & \multicolumn{1}{c}{\bf func} & \multicolumn{1}{c}{\bf cont.} & \multicolumn{1}{c}{\bf func}\\
\multicolumn{6}{l}{\bf eBay vs. Legal Onion Drugs} \\
\hline
NB & 91.4 & 57.8 & 90.5 & 56.9 & 92.2\\
SVM & 63.8 & 64.7 & 63.8 & 68.1 & 63.8\\
BoE$_\mathrm{sum}$ & 66.4 & 56.0 & 63.8 & 50.9 & 76.7\\
BoE$_\mathrm{average}$ & 75.0 & 55.2 & 59.5 & 50.0 & 75.0\\
seq2vec & 73.3 & 53.8 & 65.5 & 65.5 & 75.0\\
attention & 82.8 & 57.5 & 85.3 & 62.1 & 82.8\\
\\
\multicolumn{6}{l}{\bf Legal vs. Illegal Onion Drugs} \\
\hline
NB & 77.6 & 53.4 & 87.9 & 51.7 & 77.6\\
SVM & 63.8 & 66.4 & 63.8 & 70.7 & 63.8\\
BoE$_\mathrm{sum}$ & 52.6 & 61.2 & 74.1 & 50.9 & 51.7\\
BoE$_\mathrm{average}$ & 57.8 & 57.8 & 52.6 & 55.2 & 50.9\\
seq2vec & 56.9 & 55.0 & 54.3 & 59.5 & 49.1\\
attention & 64.7 & 51.4 & 62.9 & 55.2 & 69.0
\end{tabular}
\caption{Test accuracy in percents for each classifier (rows) in each setting (columns) on drugs-related data.
\label{tab:results_drugs}}
\end{table}

\begin{table}[t]
\centering
\setlength\tabcolsep{4pt}
\begin{tabular}{l *{5}{R}}
&& \multicolumn{1}{c}{\bf drop} & \multicolumn{1}{c}{\bf drop} & \multicolumn{1}{c}{\bf pos} & \multicolumn{1}{c}{\bf pos}\\
& \multicolumn{1}{c}{\bf full} & \multicolumn{1}{c}{\bf cont.} & \multicolumn{1}{c}{\bf func} & \multicolumn{1}{c}{\bf cont.} & \multicolumn{1}{c}{\bf func}\\
\multicolumn{6}{l}{\bf Legal vs. Illegal Onion Forums} \\
\hline
NB & 74.1 & 50.9 & 78.4 & 50.9 & 72.4\\
SVM & 85.3 & 75.9 & 56.0 & 81.9 & 81.0\\
BoE$_\mathrm{sum}$ & 25.9 & 32.8 & 21.6 & 36.2 & 35.3\\
BoE$_\mathrm{average}$ & 40.5 & 42.2 & 31.9 & 48.3 & 53.4\\
seq2vec & 50.0 & 48.9 & 50.9 & 28.4 & 51.7\\
attention & 31.0 & 37.2 & 33.6 & 27.6 & 30.2\\
\\
\multicolumn{6}{l}{\bf Trained on Drugs, Tested on Forums} \\
\hline
NB & 78.4 & 63.8 & 89.7 & 63.8 & 79.3\\
SVM & 62.1 & 69.0 & 54.3 & 69.8 & 62.1\\
BoE$_\mathrm{sum}$ & 45.7 & 50.9 & 49.1 & 50.9 & 50.0\\
BoE$_\mathrm{average}$ & 49.1 & 51.7 & 51.7 & 52.6 & 58.6\\
seq2vec & 51.7 & 61.1 & 51.7 & 54.3 & 57.8\\
attention & 65.5 & 59.2 & 65.5 & 50.9 & 66.4
\end{tabular}
\caption{Test accuracy in percents for each classifier (rows) in each setting (columns) on forums data.
\label{tab:results_forums}}
\end{table}

\subsection{Results} \label{subsec:results}

The accuracy scores for the different classifiers and settings are reported in Table~\ref{tab:results_drugs}.

\paragraph{Legal Onion vs. eBay.}

This control experiment shows that Legal Onion content is quite easily
distinguishable from eBay content, as a Naive Bayes bag-of-words classifier
reaches 91.4\% accuracy on this classification.
Moreover, replacing function words by their parts of speech even improves
performance, suggesting that the content words are the important factor in this classification.
This is confirmed by the drop in accuracy when content words are removed.
However, in this setting (drop cont.), non-trivial performance is still
obtained by the SVM classifier, suggesting that the domains are distinguishable (albeit to a lesser extent)
based on the function word distribution alone.

Surprisingly, the more sophisticated neural classifiers perform worse than Naive Bayes.
This is despite using pre-trained word embeddings,
and architectures that have proven beneficial for text classification.
It is likely that this is due to the small size of the training data, as well as the specialized
vocabulary found in this domain, which is unlikely to be supported well by the pre-trained embeddings (see \S\ref{sec:ner}).


\paragraph{Legal vs. illegal drugs.}

Classifying legal and illegal pages within the drugs domain on Onion
proved to be a more difficult task.
However, where content words are replaced with their POS tags,
the SVM classifier distinguishes between legal and illegal texts with quite a high accuracy (70.7\% on a balanced test set).
This suggests that the syntactic structure is sufficiently different between the 
domains, so as to make them distinguishable in terms of their distribution of grammatical categories.
%

\begin{figure*}[t]
\centering
\small
\begin{tabular}{l|l}
\multicolumn{1}{c}{\textbf{Legal Onion}} &
\multicolumn{1}{c}{\textbf{Illegal Onion}}\\[.1cm]
Generic Viagra Oral Jelly is used for Erectile Dys&8/03/2017 - ATTN! Looks like SIGAINT email provid\\[.1cm]
Fortis Testosteron Testosterone Enanthate 250mgml &Medical Grade Cannabis Buds We stock high quality \\[.1cm]
Generic Cialis is used to treat erection problems &1 BTC 2630.8 USD\\[.1cm]
2 Kits Misoprostol 200mg with 4 Tablets \$75 \$150&Cialis(r) is in a class of drugs called Phosphodie\\[.1cm]
Boldenone 300 New Boldenone 300 Olymp Labs&TZOLKIN CALENDAR, 140ug LSD Marquis PreTest\\[.1cm]
(generic Viagra) unbranded generic Sildenafil citr&Formed in early 2016, OzDrugsExpress is the work o\\[.1cm]
Mesterolone New Mesterolone Olymp Labs&500mg x 30 pills Price:$ 45.95 Per pill:$ 1.53 Ord\\[.1cm]
manufactured by: Cipla Pharmaceuticals Compare to &Generic Kaletra contains a combination of lopinavi\\[.1cm]
(generic Zoloft) Sertraline 25 mg tablets US\$31.05&/ roduct-category/cannabis/ Cannabis / 14g Amnesia\\[.1cm]
Sustanon-250mg Testosterone Compound 250mgml&Welcome to SnowKings Good Quality Cocaine[.1cm]
\end{tabular}
\caption{Example paragraphs (data instances) from the training sets of the Legal Onion and Illegal Onion subsets of the drug-related corpus (ten examples from each).
Each paragraph is trimmed to the first 50 characters for space reasons.
Distinguishing lexical features are easily observed, e.g.,
names of legal and illegal drugs.
\label{fig:examples}}
\end{figure*}

\begin{figure*}[t]
\centering
\small
\begin{tabular}{l|l}
\multicolumn{1}{c}{\textbf{Legal Onion}} &
\multicolumn{1}{c}{\textbf{Illegal Onion}}\\[.1cm]
( ADJ PROPN ) PROPN PROPN VERB NUM &3. We VERB VERB NOUN with ADV the A\\[.1cm]
( ADJ PROPN , PROPN ) PROPN NUM NOU&NOUN . NOUN NOUN PROPN NOUN\\[.1cm]
PROPN PROPN VERB PROPN NUM NOUN US\$&VERB NOUN NOUN : No ADJ than NUM NO\\[.1cm]
PROPN PROPN PROPN NUM NOUN \$ NUM PR&Welcome! We VERB ADJ to VERB a ADJ \\[.1cm]
NOUN NUM PROPN with NOUN - NOUN NUM&PROPN PROPN PROPN NOUN : ADJ NOUN V\\[.1cm]
PROPN NUM PROPN PROPN \$ NUM&( PROPN ) NUM PROPN PROPN NUM PROPN\\[.1cm]
ADJ PROPN NUM PROPN VERB US\$ NUM fo&NOUN PROPN NUM , NUM ADJ NOUN , ADJ\\[.1cm]
( ADJ PROPN ) NOUN NUM NOUN NOUN . &You VERB ADV VERB NUM of these NOUN\\[.1cm]
PROPN PROPN PROPN NUM PROPN PROPN -&/ NOUN - NOUN / NOUN PROPN / PROPN \\[.1cm]
PROPN / PROPN PROPN PROPN / PROPN P&Any NOUN VERB us.\\[.1cm]
\end{tabular}
\caption{Example paragraphs (data instances) from the training sets of the Legal Onion and
Illegal Onion subsets of the drug-related corpus (ten examples from each),
where content words are replaced with their parts of speech.
Each paragraph is trimmed to the first 50 characters for space reasons.
Different instances are shown than in Figure~\ref{fig:examples}.
Although harder to identify, distinguishing patterns are observable in this case too.
\label{fig:examples_poscontent}}
\vspace{-.2cm}
\end{figure*}

\section{Illegality Detection Across Domains} \label{sec:cross_domains}

To investigate illegality detection across different domains, 
we perform classification experiments on the  ``forums'' category that is also separated into legal and illegal sub-categories in DUTA-10K.
The forums  contain user-written text in various topics. Legal forums often discuss web design and other technical
and non-technical activity on the internet, while illegal ones involve
discussions about cyber-crimes and guides on how to commit them,
as well as narcotics, racism and other criminal activities.
As this domain contains user-generated content, it is more varied
and noisy.

\subsection{Experimental setup}
We use the cleaning process described in Section~\ref{sec:datasets} and data splitting described in Section~\ref{sec:classification}, with the same number of paragraphs.
We experiment with two settings:
\begin{itemize}
\item Training and testing on Onion legal vs. illegal forums,
to evaluate whether the insights observed in the drugs domain generalize to user-generated content.
\item Training on Onion legal vs. illegal drugs-related pages,
and testing on Onion legal vs. illegal forums.
This cross-domain evaluation reveals whether the distinctions learned on the
drugs domain generalize directly to the forums domain.
\end{itemize}  

\subsection{Results}

Accuracy scores are reported in Table~\ref{tab:results_forums}.

\paragraph{Legal vs. illegal forums.}

Results when training and testing on forums data are much worse for the neural-based systems,
probably due to the much noisier and more varied nature of the data. 
However, the SVM model achieves an accuracy of 85.3\% in the full setting. 
Good performance is presented by this model even in the cases where the content words are dropped (drop. cont.) 
or replaced by part-of-speech tags (pos cont.),
underscoring the distinguishability of legal in illegal content based on shallow syntactic structure in this domain as well.

\paragraph{Cross-domain evaluation.}

Surprisingly, training on drugs data and evaluating on forums performs much
better than 
in the in-domain setting
for four out of five classifiers.
This implies that while the forums data is noisy, it can be accurately classified into legal and illegal content
when training on the cleaner drugs data. This also shows that illegal texts in Tor share common properties regardless of topical category.
The much lower results obtained by the models where content words are dropped (drop cont.) or converted to POS tags (pos cont.),
namely less than 70\% as opposed to 89.7\% when function words are dropped, suggest that some of these properties are lexical.

\begin{table}[t]
\small
\centering
\begin{tabular}{p{21mm}r|}
\multicolumn{2}{c}{\textbf{Legal Onion}}\\
feature & ratio\\
\hline
cart & 0.037\\
2016 & 0.063\\
Bh & 0.067\\
drugs & 0.067\\
EUR & 0.077\\
very & 0.083\\
Per & 0.091\\
delivery & 0.091\\
symptoms & 0.091\\
Please & 0.100\\
Quantity & 0.100\\
here & 0.100\\
check & 0.100\\
contact & 0.100\\
called & 0.100\\
not & 0.102\\
if & 0.105\\
If & 0.111\\
taking & 0.111\\
like & 0.111\\
questions & 0.111\\
Cart & 0.118\\
$>$ & 0.125\\
take & 0.125\\
\end{tabular}
\begin{tabular}{p{21mm}r}
\multicolumn{2}{c}{\textbf{Illegal Onion}}\\
feature & ratio\\
\hline
Laboratories & 8.500\\
Cipla & 9.000\\
Pfizer & 9.000\\
each & 9.571\\
Force & 10.000\\
Pharma & 10.333\\
grams & 10.333\\
300 & 10.500\\
Bitcoin & 11.500\\
RSM & 12.000\\
Sun & 12.000\\
manufactured & 12.667\\
Tablets & 13.000\\
tablets & 13.714\\
120 & 16.000\\
Moldova & 17.000\\
citrate & 18.000\\
Pharmaceuticals & 31.500\\
New & 33.000\\
200 & 36.000\\
blister & 48.000\\
generic & 49.500\\
60 & 55.000\\
@ & 63.000\\
\end{tabular}
\caption{Most indicative features for each class in NB classifier
trained on Onion legal vs. illegal drugs,
and the ratio between their number of occurrences in the ``illegal''
class and the ``legal'' class in the training set.
Left: features with lowest ratio; right: features with highest ratio.
While some strong features are entities, many are in fact function words.
\label{tab:nb_weights}}
\end{table}

\section{Discussion} \label{sec:discussion}

  As shown in Section~\ref{sec:domain}, the Legal Onion and Illegal Onion domains
  are quite distant in terms of word distribution and named entity Wikification.
  Moreover, named entity recognition and Wikification work less well
  for the illegal domain, and so do state-of-the-art neural text classification architectures (Section~\ref{sec:classification}),
  which present inferior results to simple bag-of-words model. This is likely a result of the different vocabulary and syntax
  of text from Onion domain, compared to standard domains used for training NLP models and pre-trained word embeddings.
  This conclusion has practical implications: to effectively process text in Onion, considerable domain adaptation should be performed,
  and effort should be made to annotate data and extend standard knowledge bases to cover this idiosyncratic domain.

  Another conclusion from the classification experiments is that
  the Onion Legal and Illegal Onion texts are harder to distinguish than
  eBay and Legal Onion, meaning that deciding on domain boundaries should consider syntactic structure, and not only lexical differences.

  \paragraph{Analysis of texts from the datasets.}

    Looking at specific sentences (Figure~\ref{fig:examples})
    reveals that Legal Onion and Illegal Onion are easy to distinguish
    based on the identity of certain words, e.g., terms for legal and illegal drugs,
    respectively.
    Thus looking at the word forms is already a good solution for tackling this
    classification problem,
    which gives further insight as to why modern text classification (e.g., neural networks)
    do not present an advantage in     terms of accuracy.

  \paragraph{Analysis of manipulated texts.}

    Given that replacing content words with their POS tags substantially lowers
    performance for classification of legal vs illegal drug-related texts
    (see ``pos cont.'' in Section~\ref{sec:classification}),
    we conclude that the distribution of parts of speech alone is not as strong
    a signal as the word forms for distinguishing between the domains.
    However, the SVM model 
    does manage to distinguish between the texts even in this setting.
    Indeed, Figure~\ref{fig:examples_poscontent} demonstrates that
    there are easily identifiable patterns distinguishing between the domains,
    but that a bag-of-words approach may not be sufficiently expressive to identify them. 

  \paragraph{Analysis of learned feature weights.}

    As the Naive Bayes classifier was the most successful at distinguishing
    legal from illegal texts in the full setting (without input manipulation),
    we may conclude that the very occurrence of certain words provides a strong indication
    that an instance is taken from one class or the other.
    Table~\ref{tab:nb_weights} shows the most indicative features learned
    by the Naive Bayes classifier for the
    Legal Onion vs. Illegal Onion classification in this setting.
    Interestingly, many strong features are function words,
    providing another indication of the different distribution of function words in the two domains.

\section{Conclusion} \label{sec:conclusion}

  In this paper we identified several distinguishing factors between legal and illegal texts, taking
  a variety of approaches, predictive (text classification), application-based (named entity Wikification),
  as well as an approach based on raw statistics. Our results revealed that legal and illegal texts on the Darknet
  are not only distinguishable in terms of their words, but also in terms of their shallow syntactic structure, manifested
  in their POS tag and function word distributions. Distinguishing features between legal and illegal texts are
  consistent enough between domains, so that a classifier trained on drug-related websites can be straightforwardly
  ported to classify legal and illegal texts from another Darknet domain (forums).
  Our results also show that in terms of vocabulary, legal texts and illegal texts are as distant from each other, 
  as from comparable texts from eBay.
 
  We conclude from this investigation that Onion pages provide an attractive testbed for studying distinguishing
    factors between the text of legal and illegal webpages, as they present challenges to off-the-shelf NLP tools,
    but at the same time have sufficient self-consistency to allow studies of the linguistic signals that separate these classes.

\section*{Acknowledgments}

This work was supported by the Israel Science Foundation (grant no. 929/17) and by the HUJI Cyber Security Research Center in conjunction with the Israel National Cyber Bureau in the Prime Minister's Office. We thank Wesam Al Nabki, Eduardo Fidalgo, Enrique Alegre, and Ivan de Paz for sharing their dataset of Onion sites with us.

\bibliography{acl2019}
\bibliographystyle{acl_natbib}

\end{document}